\documentclass[letterpaper]{article} 
\usepackage{times}  
\usepackage{helvet}  
\usepackage{courier}  
\usepackage{url}  
\usepackage{graphicx}  
\usepackage{amsmath}
\usepackage{amssymb}
\usepackage{algorithm}
\usepackage{algpseudocode}
\usepackage{subfig}

\begin{document}
%
\title{Online Evaluation of Audiences for Targeted Advertising via Bandit Experiments\footnote{The authors are part of \texttt{JD Intelligent Ads Lab}. The views represent that of the authors, and not \texttt{JD.com}. We thank Jun Hao, Lei Wu and Paul Yan for their helpful collaboration, and Caio Waisman for extensive comments. Previous version: July 5, 2019.}}
\author{Tong Geng, \texttt{JD.com} \\tong.geng@jd.com\\\\Xiliang Lin, \texttt{JD.com}\\ xiliang.lin@jd.com\\\\ Harikesh S. Nair, {\texttt{JD.com} and Stanford University}\\harikesh.nair@stanford.edu}

\maketitle
\begin{abstract}
Firms implementing digital advertising campaigns face a complex problem in determining the right match between their advertising creatives and target audiences. Typical solutions to the problem have leveraged non-experimental methods, or used ``split-testing" strategies that have not explicitly addressed the complexities induced by targeted audiences that can potentially overlap with one another. This paper presents an adaptive algorithm that addresses the problem via online experimentation. The algorithm is set up as a contextual bandit and addresses the overlap issue by partitioning the target audiences into disjoint, non-overlapping sub-populations. It learns an optimal creative display policy in the disjoint space, while assessing in parallel which creative has the best match in the space of possibly overlapping target audiences. Experiments show that the proposed method is more efficient compared to naive ``split-testing'' or non-adaptive ``A/B/n'' testing based methods. We also describe a testing product we built that uses the algorithm. The product is currently deployed on the advertising platform of \texttt{JD.com}, an eCommerce company and a publisher of digital ads in China. 
\end{abstract}

\section{Introduction}

A critical determinant of the success of advertising campaigns is picking the right audience to target. As digital ad-markets have matured and the ability to target advertising has improved, the range of targeting options has expanded, and the profile of possible audiences have become complex. Both advertisers and publishers now rely on data-driven methods to evaluate audiences and to find effective options with which to advertise to them. This paper presents a new bandit algorithm along with a product built to facilitate such evaluations via online experimentation.

The problem addressed is as follows. An advertiser designing a campaign wants to pick, from a set of $\mathbb{K} = \{1,..,K\}$ possible target audiences and $\mathbb{R} = \{1,..,R\}$ creatives, a combination $k,r$ ($k\in\mathbb{K}$, $r\in\mathbb{R}$) that provides her the highest expected payoff. The target audiences can be complex, potentially overlapping with each other, and the creatives can be any type of media (picture, video, text etc). We would like to design an experiment to find the best creative-target audience combination while minimizing the costs of experimentation to the advertiser.

Consider an archetypal experimental design in which each creative-target audience combination forms a test arm, so that the goal of the test is to discover the arm with the highest expected payoff. To implement such a design, we need to address two challenges associated this problem.

The first difficulty is the possibility of overlap in target audiences that are being compared (e.g., ``San Francisco users'' and ``Male users''). This generates a complication in user assignment in the test because it is not obvious to which constituent arm, a user belonging to an overlapping region should be assigned (e.g., should a Male user from San Francisco be assigned to the ``San Francisco-creative'' arm or the ``Male-creative'' arm?). Assigning the overlapping user to one of the constituent arms violates the representativeness of the arms (e.g., if we use a rule that Male users from San Francisco will always be assigned to the ``San Francisco-creative'' arm, the ``Male-creative'' arm will have no San Franciscans, and will not represent the distribution of Male users in the platform population). Such assignment also under-utilizes data: though the feedback from the user is informative of all constituent arms, it is being used to learn the best creative for only one picked arm (e.g., if we assign a Male user from San Francisco to the ``San Francisco-creative'' arm, we do not learn from him the value of the ``Male-creative'' arm, even though his behavior is informative of that arm).

The second difficulty is that typical ``A/B/n'' test designs keep the sample/traffic splits constant as the test progresses. Therefore, both good and bad creatives will be allocated the same amount of traffic during the test. Instead, as we learn during the test that an arm is not performing well, reducing its traffic allocation can reduce the cost to the advertiser of experimentation.

The goal of this paper is to develop an algorithm that addresses both issues. It has two broad steps. In step one, we split the compared target audiences (henceforth ``\textit{TA}''s) into disjoint audience sub-populations (henceforth ``\textit{DA}"s), so that the set of \emph{DA}s fully span the set of \emph{TA}s. In step two, we train a bandit with the creatives as arms, the payoffs to the advertiser as rewards, and the \emph{DA}s, rather than the \emph{TA}s as the contexts. As the test progresses, we aggregate over all \emph{DA}s that correspond to each \emph{TA} to adaptively learn the best creative-\emph{TA} match. In essence, we learn an optimal creative allocation policy at the disjoint sub-population level, while making progress towards the test goal at the \emph{TA} level. Because the \emph{DA}s have no overlap, each user can be mapped to a distinct \emph{DA}, addressing the assignment problem. Because all \emph{DA}s that map to a \emph{TA} help inform the value of that \emph{TA}, learning is also efficient. Further, tailoring the bandit's policy to a more finely specified context $-$ i.e., the \emph{DA} $-$ allows it to match the creative to the user's tastes more finely, thereby improving payoffs and reducing expected regret, while delivering on the goal of assessing the best combination at the level of a more aggregated audience. The adaptive nature of the test ensures the traffic is allocated in a way that reduces the cost to the advertiser from running the test, because creatives that are learned to have low value early are allocated lesser traffic within each \emph{DA} as the test progresses. The overall algorithm is implemented as a contextual Thompson Sampler (henceforth ``TS''; see \cite{russo2018} for an overview).

Increasing the overlap in the tested \emph{TA}s increases the payoff similarity between the \emph{TA}s, making it harder to detect separation. One worry is that the TS in such situations requires enormous amounts of data before stopping, and performance is degraded to the extent that it is practically unviable. An attractive feature of the proposed algorithm is that feedback on the performance of \emph{DA}s helps inform the performance of all \emph{TA}s to which they belong. This \emph{cross-audience learning} serves as a counterbalancing force that keeps performance stable as overlap increases, preventing the sample sizes required to stop the test from growing unacceptably large and making the algorithm impractical.

In several simulations, we show the proposed TS performs well in realistic situations, including with high levels of overlap; and is competitive against benchmark methods including non-adaptive designs and ``split-testing" designs currently used in industry. To illustrate real-world performance, we also discuss a case-study from a testing product on the advertising platform of \texttt{JD.com}, where the algorithm is currently deployed.

\section{Related Work and Other Approaches}{\label{sec:lit-review}}

There is a mature literature on successful applications of bandits in web content optimization (e.g., \cite{AgarwalChenElango2009}, \cite{LiChuLangfordSchapire2010}, \cite{ChapelleLi2011}, \cite{HauseretalBannerMorphing2015}, \cite{AgarwaletalBanditTechDebt16}). This paper belongs to a sub-stream of this work that has focused on using bandits for controlled experiments on the web. The closest papers to our work are the complementary papers by \cite{scott2015multi}, \cite{schwartzetal2017} and \cite{Juetal2019} who propose using bandit experiments to evaluate creatives for targeted advertising, without focusing explicitly on the problem addressed here of comparing target audiences.

In industry, the popular experimental design to compare \emph{TA}s for advertising campaigns is sometimes called ``audience split-testing" (e.g., \cite{facebooksplit2019}, \cite{tencentsplit2019}). Suppose there is only one creative, and $K$ \emph{TA}s are to be compared. The audience split-testing design randomizes test users into $K$ arms, each of which is associated with the same creative, but which correspond respectively to the \emph{K} \emph{TA}s. Conditional on being randomized into an arm, a user is shown the creative only if his features match the arms' \emph{TA} definition. This ensures that the mix of overlapping and non-overlapping audiences is representative; however, the design under-utilizes the informational content of experimental traffic as there is no learning from users who are randomized into a test-arm but do not match its \emph{TA} definition. Also, in contrast to the design proposed here, there is no cross-audience learning from overlapping users. In addition, the typical implementation of split-testing is non-adaptive, and is not cost minimizing unlike the adaptive design presented here.

A possible strategy for maintaining the representativeness of \emph{TA}s in the test is to randomly allocate some proportion $p$ of users in each overlapping region to the \emph{TA}s the region overlaps with. Unfortunately, no value of $p$ exists that maintains representativeness after such allocation while retaining all the data. To illustrate, suppose we have two \emph{TA}s ($TA1$ and $TA2$) that overlap with each other, so we have three \emph{DA}s, $DA1$, $DA2$ and $DA3$, with $DA2$ belonging to both $TA1$ and $TA2$. Suppose in the test, a representative sample of $N_{DA1}$, $N_{DA2}$, and $N_{DA3}$ users belonging to each of the three \emph{DA}s arrive, and have to be assigned in this manner to $TA1$ and $TA2$. If we allocate proportion $p$ of users in $DA2$ to $TA1$, the proportion of $DA2$ users in $TA1$ is $P(DA2|TA1)=\frac{p \times N_{DA2}}{p \times N_{DA2} + N_{DA1}}$. However, to be representative of the population, we need this proportion to be $\frac{N_{DA2}}{N_{DA2} + N_{DA1}}$. The only value of $p$ that makes $TA1$ under this scheme representative is 1. However, when $p=1$, the proportion of $DA2$ in $TA2$ is 0, making $TA2$ under this scheme not representative of $TA2$ in the population. One can restore representativeness by dropping a randomly picked proportion $1-p$ of $N_{DA1}$ users and $p$ of $N_{DA2}$ users. But this involves throwing away data and induces the same issue as the ``audience split-testing" design above of under-utilizing the informational content of experimental traffic.

\section{Method}

\subsection{Step 1: Setup}

We take as input into the test the $\mathbb{K} = \{1,..,K\}$ possible \emph{TA}s and $\mathbb{R} = \{1,..,R\}$ creatives the advertiser wants to compare. In step 1, we  partition the users in the $K$ \emph{TA}s into a set $\mathbb{J} = \{1,..,J\}$ of $J$ \emph{DA}s. For example, if the \emph{TA}s are ``San Francisco users'' and ``Male users,'' we create three \emph{DA}s, ``San Francisco users, Male,'' ``San Francisco users, Not Male," and ``Non San Francisco users, Male.''

\subsection{Step 2: Contextual Bandit Formulation}

In step 2, we treat each \emph{DA} as a context, and each creative as an arm that is pulled adaptively based on the context. When a user $i$ arrives at the platform, we categorize the user to a context based on his features, i.e.,
\begin{equation}
    i\in DA(j) \text{ if } i\text{'s features match the definition of } j,
\end{equation}
where $DA(j)$ denotes the set of users in DA $j$. A creative $r\in\mathbb{R}$ is then displayed to the user based on the context. The cost of displaying creative $r$ to user $i$ in context $j$ is denoted as $b_{irj}$. After the creative is displayed, the user's action, $y_{irj}$, is observed. The empirical implementation of the product uses clicks as the user feedback for updating the bandit, so $y$ is treated as binary, i.e., $y_{irj}\in\{0,1\}$. The payoff to the advertiser from the ad-impression, $\pi_{irj}$, is defined as:
\begin{equation}
    \pi_{irj}=\gamma\cdot y_{irj}-b_{irj},    
\end{equation}
where $\gamma$ is a factor that converts the user's action to monetary units. The goal of the bandit is to find an optimal policy $g(j):\mathbb{J}\rightarrow\mathbb{R}$ which allocates the creative with the maximum expected payoff to a user with context $j$.

\subsubsection{Thompson Sampler}

To develop the TS, we model the outcome $y_{irj}$ in a Bayesian framework, and let
\begin{eqnarray}
y_{irj}  \sim  p(y_{irj}|\theta_{rj}),\\
\theta_{rj}  \sim p(\theta_{rj}|\Omega_{rj}).
\end{eqnarray}
where $\theta_{rj}$ are the parameters that describe the distribution of action $y_{irj}$, and $\Omega_{rj}$ are the hyper-parameters governing the distribution of $\theta_{rj}$. Since $y$ is Bernoulli distributed, we make the typical assumption that the prior on $\theta$ is Beta which is conjugate to the Bernoulli distribution. With $\Omega_{rj}\equiv(\alpha_{rj},\beta_{rj})$, we model,
\begin{eqnarray}
y_{irj}\sim \texttt{Ber}(\theta_{rj}),\\
\theta_{rj}\sim\texttt{Beta}(\alpha_{rj},\beta_{rj}).
\end{eqnarray}
Given $y_{irj}\sim \texttt{Ber}(\theta_{rj})$, the expected payoff of each creative-disjoint sub-population combination (henceforth ``\emph{C-DA}'') is:
\begin{equation}
    \mu_{rj}^{\pi}(\theta_{rj})=\mathbb{E}[\pi_{irj}]=\gamma\mathbb{E}[y_{irj}]-\mathbb{E}[b_{irj}]\\=\gamma\theta_{rj}-\bar{b}_{rj},\forall r\in\mathbb{R},j\in\mathbb{J},
\end{equation}
where $\bar{b}_{rj}$ is the average cost of showing the creative $r$ to the users in $DA(j)$.\footnote{$\gamma$ may be determined from prior estimation or advertisers' judgment of the value attached to users' actions. $\gamma$ is pre-computed and held fixed during the test. $\bar{b}_{rj}$ and $\hat{p}(j|k)$ (defined later) can be pre-computed outside of the test from historical data and held fixed during the test, or inferred during the test using a simple bin estimator that computes these as averages over the observed cost and user contexts data.} 

To make clear how the bandit updates parameters, we add the index $t$ for batch. Before the test starts, $t=1$, we set diffuse priors and let $\alpha_{rj,t=1}=1,\beta_{rj,t=1}=1,\forall r \in \mathbb{R}, j \in \mathbb{J}$. This prior implies the probability of taking action $y$, $\theta_{rj,t=1},\forall r \in \mathbb{R}, j \in \mathbb{J}$ is uniformly distributed between 0\% and 100\%.

In batch $t$, $N_{t}$ users arrive. The TS displays creatives to these users dynamically, by allocating each creative according to the posterior probability each creative offers the highest expected payoffs given the user's context. Given the posterior at the beginning of batch $t$, the probability a creative $r$ provides the highest expected payoff is,
\begin{equation*}
    w_{rjt}=
    Pr[\mu_{rj}^{\pi}(\theta_{rjt})=\max\limits_{r \in \mathbb{R}} (\mu_{rj}^{\pi}(\theta_{rjt}))|\vec{\alpha}_{jt},\vec{\beta}_{jt}],
\end{equation*}
where $\vec{\alpha}_{jt} = [\alpha_{1jt}, \dots, \alpha_{Rjt}]'$ and $\vec{\beta}_{jt} = [\beta_{1jt}, \dots, \beta_{Rjt}]'$ are the parameters of the posterior distribution of $\vec{\theta}_{jt} = [\theta_{1jt},\dots,\theta_{Rjt}]'$.

To implement this allocation, for each user $i=1,..,N_{t}$ who arrives in batch $t$, we determine his context $j$, and make a draw of the $R\times1$ vector of parameters, $\tilde{\boldsymbol{\theta}}_{jt}^{\left(i\right)}$. Element $\tilde{\theta}_{rjt}^{\left(i\right)}$ of the vector is drawn from $\texttt{Beta}(\alpha_{rjt}, \beta_{rjt})$ for $r\in\mathbb{R}$. Then, we compute the payoff for each creative $r$ as $\mu_{rj}^{\pi}(\tilde{\theta}_{rjt}^{\left(i\right)})=\gamma\tilde{\theta}_{rjt}^{\left(i\right)}-\bar{b}_{rj}$, and display to $i$ the creative with the highest $\mu_{rj}^{\pi}(\tilde{\theta}_{rjt}^{\left(i\right)})$.

We update all parameters at the end of processing the batch, after the outcomes for all users in the batch is observed. We compute the sum of binary outcomes for each \emph{C-DA} combination as,
\begin{equation}
    s_{rjt}=\sum_{i=1}^{n_{rjt}}y_{irjt}, \forall r \in \mathbb{R},j \in \mathbb{J},
\end{equation}
where $n_{rjt}$ is the number of users with context $j$ allocated to creative $r$ in batch $t$. Then, we update parameters as:
\begin{equation*}
     \vec{\alpha}_{j(t+1)}=\vec{\alpha}_{jt}+\vec{s}_{jt}, \vec{\beta}_{j(t+1)}=\vec{\beta}_{jt}+\vec{n}_{jt}-\vec{s}_{jt}, \forall j \in \mathbb{J},
\end{equation*}
where $\vec{s}_{jt} = [s_{1jt},\dots,s_{Rjt}]'$, and $\vec{n}_{jt} = [n_{1jt},\dots,n_{Rjt}]'$.

Then, we enter batch $t+1$, and use $\vec{\alpha}_{j(t+1)}$ and $\vec{\beta}_{j(t+1)}$ as the posterior parameters to allocate creatives at $t+1$. We repeat this process until a pre-specified stopping condition (outlined below) is met.

\subsubsection{Probabilistic Aggregation and Stopping Rule}
While the  contextual bandit is set up to learn the best \emph{C-DA} combination, the goal of the test is to learn the best creative-target audience combination (henceforth ``\emph{C-TA}''). As such, we compute the expected payoff of each \emph{C-TA} combination by aggregating the payoffs of corresponding \emph{C-DA} combinations, and stop on the basis of the regret associated with learning the best \emph{C-TA} combination.

Using the law of total probability, we can aggregate across all \emph{C-DA}s associated with \emph{C-TA} combination $(r,k)$ to obtain $\lambda_{rkt}$,
\begin{equation}
    \lambda_{rkt}=\sum_{j\in \mathcal{O}(k)}\theta_{rjt}\cdot\hat{p}(j|k).
    \label{eq:lambda}
\end{equation}
In equation (\ref{eq:lambda}), $\lambda_{rkt}$ is the probability that a user picked at random \textit{from within TA($k$)} in batch $t$, takes the action $y=1$ upon being displayed creative $r$; $\hat{p}(j|k)$ is the probability (in the platform population) that a user belonging to $TA(k)$ is also of the context $j$; and $\mathcal{O}(k)$ is the set of disjoint sub-populations ($j$s) whose associated \emph{DA($j$)}s are subsets of $TA(k)$.

Given equation (\ref{eq:lambda}), the posterior distribution of $\theta_{rjt}$s from the TS induces a distribution of $\lambda_{rkt}$s. We can obtain draws from this distribution using Monte Carlo sampling. For each draw $\ensuremath{\theta_{rkt}^{\left(h\right)}},h=1,..,H$ from $\texttt{Beta}(\alpha_{rjt}, \beta_{rjt})$, we can use equation (\ref{eq:lambda}) to construct a corresponding $\ensuremath{\lambda_{rkt}^{\left(h\right)}},h=1,..,H$. For each such $\lambda_{rkt}^{\left(h\right)}$,  we can similarly compute the implied expected payoff to the advertiser from displaying creative $r$ to a user picked at random from within TA($k$) in batch $t$,
\begin{equation}
\omega_{rkt}^{\pi}(\lambda_{rk}^{\left(h\right)})=\gamma\lambda_{rkt}^{\left(h\right)}-\bar{b}_{rk},\forall r\in\mathbb{R},k\in\mathbb{K},
    \label{eq:omega}
\end{equation}
where $\bar{b}_{rk}$ is the average cost for showing creative $r$ to target audience $k$, which can be obtained by aggregating $\bar{b}_{rj}$ through analogously applying equation (\ref{eq:lambda}).
Taking the $H$ values of $\omega_{rkt}^{\pi}(\lambda_{rk}^{\left(h\right)})$ for each $(r,k)$, we let $r_{kt}^{*}$ denote the creative that has the highest expected payoff within each \emph{TA} $k$ across all $H$ draws, i.e., 
\begin{equation}
r_{kt}^{*}=\underset{r\in\mathbb{R}}{arg\max}\underset{h=1,..,H}{\max}\omega_{rkt}^{\pi}(\lambda_{rk}^{\left(h\right)}).
\label{eq:bestr}
\end{equation}
Hence, $\omega_{r_{kt}^{*},kt}^{\pi}(\lambda_{rkt}^{(h)})$ denote the expected payoff for creative $r_{kt}^{*}$ evaluated at draw $h$. Also, define $\omega_{^{*}kt}^{\pi}(\lambda_{rkt}^{(h)})$ as the expected payoff for the creative assessed as the best for \textit{TA} $k$ in draw $h$ itself, i.e., 
\begin{equation}
\omega_{^{*}kt}^{\pi}(\lambda_{rkt}^{(h)})=\underset{r\in\mathbb{R}}{\max}\quad\omega_{rkt}^{\pi}(\lambda_{rk}^{\left(h\right)}),
\label{eq:bestomega}
\end{equation}
Following \cite{scott2015multi}, the value  $\omega_{^{*}kt}^{\pi}(\lambda_{rkt}^{(h)})- \omega_{r_{kt}^{*},kt}^{\pi}(\lambda_{rkt}^{(h)})$ represents an estimate of the regret in batch $t$ for \emph{TA} $k$ at draw $h$. Normalizing it by the expected payoff of the best creative across draws gives a unit-free metric of regret for each draw $h$ for each \emph{TA} $k$,
\begin{equation}
\rho_{kt}^{(h)}=\frac{\omega_{^{*}kt}^{\pi}(\lambda_{rkt}^{(h)})-\omega_{r_{kt}^{*},kt}^{\pi}(\lambda_{rkt}^{(h)})}{\omega_{r_{kt}^{*},kt}^{\pi}(\lambda_{rkt}^{(h)})},
   \label{eq:unitfreeReg}
\end{equation}

Let $pPVR(k,t)$ be the $95^{\textrm{th}}$ percentile of $\rho_{kt}^{(h)}$ across the $H$ draws. We stop the test when, 
\begin{equation}
    \max\limits_{k\in\mathbb{K}}pPVR(k,t)<0.01.
    \label{eq:pPVR}
\end{equation}
In other words, we stop the test when the normalized regret for all \textit{TA}s we are interested in falls below 0.01.\footnote{Other stopping rules may also be used, for example, based on posterior probabilities, or based on practical criteria that the test runs till the budget is exhausted (which protects the advertiser's interests since the budget is allocated to the best creative). The formal question of how to stop a TS when doing Bayesian inference is still an open issue. While data-based stopping rules are known to affect frequentist inference, Bayesian inference has traditionally been viewed as unaffected by optional stopping (e.g., \cite{edwards1963bayesian}), though the debate is still unresolved in the statistics and machine learning community (e.g., \cite{rouder} vs. \cite{dhg2018}). This paper adopts a stopping rule reflecting practical product-related considerations, and does not address this debate.} Therefore, while we learn an optimal creative displaying policy for each \emph{DA}, we stop the algorithm when we find the best creative for each \emph{TA} in terms of minimal regret. Algorithm \ref{algo:TS} shows the full procedure.

\begin{algorithm}
\caption{\emph{TS} to identify best \emph{C-TA} combination {\label{algo:TS}}}
\begin{algorithmic}[1]
\State $K$ \emph{TA}s are re-partitioned into $J$ \emph{DA}s
\State $t \gets 1$
\State $\alpha_{rjt} \gets 1,\beta_{rjt} \gets 1,\forall r \in \mathbb{R}, j \in \mathbb{J}$
\State Obtain from historical data $\hat{p}(j|k),\gamma, \bar{b}_{rj},\forall r \in \mathbb{R}, j \in \mathbb{J},  k \in \mathbb{K}$
\State $pPVR(k,t) \gets 1,\forall k \in \mathbb{K}$
\While{$\max\limits_{k\in\mathbb{K}}pPVR(k,t)<0.01$} 
\State A batch of $N_{t}$ users arrive
\ForAll{users}
\State Sample $\tilde{\theta}_{rjt}^{(i)}$ using $\texttt{Beta}(\alpha_{rjt},\beta_{rjt})$ for each $r\in\mathbb{R}$
\State Feed creative $I_{it} = argmax_{r\in \mathbb{R}} \gamma\tilde{\theta}_{rjt}^{\left(i\right)}-\bar{b}_{rjt}$
\EndFor
\State Collect data $\{y_{irjt}\}_{i=1}^{N_{t}},\{n_{rjt}\}_{r \in \mathbb{R}, j \in \mathbb{J}}$
\State Compute $s_{rjt}=\sum_{i=1}^{n_{rjt}}y_{irjt}, \forall r \in \mathbb{R},j \in \mathbb{J}$
\State Update $\alpha_{rj(t+1)}=\alpha_{rjt}+s_{rjt}, \forall r \in \mathbb{R},j \in \mathbb{J}$
\State Update $\beta_{rj(t+1)}=\beta_{rjt}+n_{rjt}-s_{rjt}, \forall r \in \mathbb{R},j \in \mathbb{J}$
\State Make $h=1,..,H$ draws of $\theta_{rj(t+1)}$s, i.e.
\begin{equation*}
\resizebox{0.8\hsize}{!}{
    $\begin{bmatrix}\begin{array}{c}
    \theta_{11(t+1)}\\
    ...\\
    \theta_{rj(t+1)}\\
    ...\\
    \theta_{RJ(t+1)}
    \end{array}\end{bmatrix}^{(h)}\sim\begin{bmatrix}
    \texttt{Beta}(\alpha_{11(t+1)},\beta_{11(t+1)})\\
    ...\\
    \texttt{Beta}(\alpha_{rj(t+1)},\beta_{rj(t+1)})\\
    ...\\
    \texttt{Beta}(\alpha_{RJ(t+1)},\beta_{RJ(t+1)})
    \end{bmatrix}^{(h)},
    \forall h=1,...,H$
}
\end{equation*}
\State Compute $\vec{\lambda}_{t+1}^{(h)}=$
\begin{equation*}
\resizebox{0.8\hsize}{!}{
    $\begin{bmatrix}\begin{array}{c}
    \lambda_{11(t+1)}\\
    ...\\
    \lambda_{rk(t+1)}\\
    ...\\
    \lambda_{RK(t+1)}
    \end{array}\end{bmatrix}^{(h)}=\begin{bmatrix}\sum\limits_{j\in O(k=1)}\hat{p}(j|k=1)\cdot\theta_{rj(t+1)}\\
    ...\\
    \sum\limits_{j\in O(k)}\hat{p}(j|k)\cdot\theta_{rj(t+1)}\\
    ...\\
    \sum\limits_{j\in O(k=K)}\hat{p}(j|k=K)\cdot\theta_{rj(t+1)}
    \end{bmatrix}^{(h)},
    \forall h=1,...,H$
}
\end{equation*}
\State Compute $\vec{\omega^{\pi}}_{t+1}^{(h)}(\vec{\lambda}_{t+1}^{(h)})=$
\begin{equation*}
\resizebox{0.8\hsize}{!}{
    $\begin{bmatrix}\begin{array}{c}
    \omega_{11(t+1)}^{\pi}\\
    ...\\
    \omega_{rk(t+1)}^{\pi}\\
    ...\\
    \omega_{RK(t+1)}^{\pi}
    \end{array}\end{bmatrix}^{(h)}=\begin{bmatrix}\gamma\cdot\lambda_{11(t+1)}-\bar{b}_{11(t+1)}\\
    ...\\
    \gamma\cdot\lambda_{rkt}-\bar{b}_{rk(t+1)}\\
    ...\\
    \gamma\cdot\lambda_{RKt}-\bar{b}_{RK(t+1)}
    \end{bmatrix}^{(h)},
    \forall h=1,...,H$
}
\end{equation*}
\State Compute $\rho_{k(t+1)}^{(h)}=\frac{\omega_{^{*}k(t+1)}^{\pi}(\lambda_{rk(t+1)}^{(h)})-\omega_{r_{k(t+1)}^{*},k(t+1)}^{\pi}(\lambda_{rk(t+1)}^{(h)})}{\omega_{r_{k(t+1)}^{*},k(t+1)}^{\pi}(\lambda_{rk(t+1)}^{(h)})}, \quad \forall h=1,...,H, k \in \mathbb{K}$
\State $\forall k \in \mathbb{K}$, calculate $pPVR(k,t+1)$ as the $95^{\textrm{th}}$ percentile across the $H$ draws of $\rho^{(h)}_{k(t+1)}$
\State Set $t \gets t+1$
\EndWhile
\end{algorithmic}
\end{algorithm}

\section{Experiments}
This section reports on experiments that establish the face validity of the TS; compares it to audience split testing and a random allocation schema where each creative is allocated to each context with equal probability; and explores its performance when the degree of overlap in \emph{TA}s increases.

\subsection{Setup}
\begin{figure}
\centering
\includegraphics[width=0.75\textwidth]{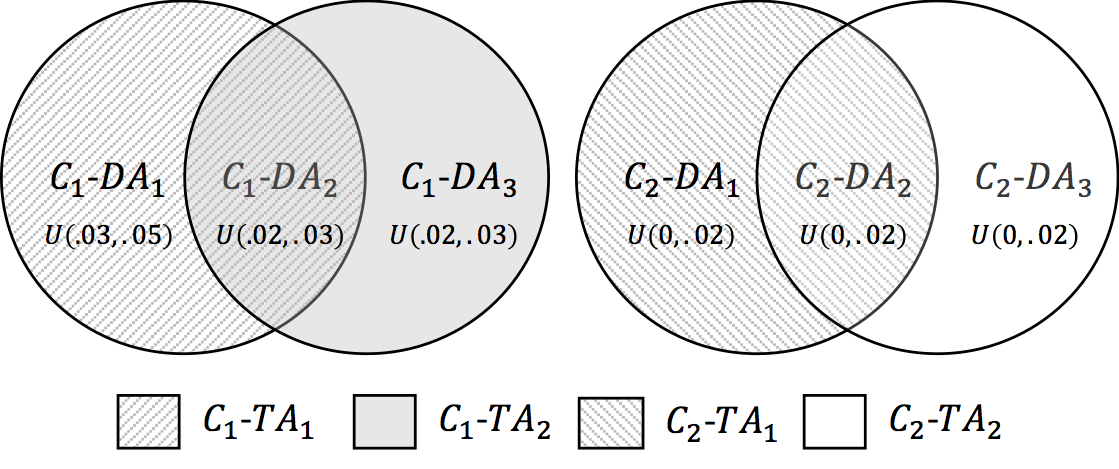}
\caption{Simulation Setup: 2 Cs, 2 TAs and 3 DAs}\label{fig:simul-setup}
\end{figure}

For the experiments, we consider a setup with 2 creatives and 2 overlapping \emph{TA}s, implying 3 \emph{DA}s, 4 \emph{C-TA} combinations and 6 \emph{C-DA} combinations as shown in Figure (\ref{fig:simul-setup}). The \emph{TA}s are assumed to be of equal sizes, with an overlap of 50\%.\footnote{Specifically, $\Pr\left(TA_{1}\right)=\Pr\left(TA_{2}\right)=.5$;  $\Pr\left(DA_{1}|TA_{1}\right)=\Pr\left(DA_{2}|TA_{1}\right)=0.5$; $\Pr\left(DA_{2}|TA_{2}\right)=\Pr\left(DA_{3}|TA_{2}\right)=0.5$; and $\Pr\left(DA_{1}|TA_{2}\right)=\Pr\left(DA_{3}|TA_{1}\right)=0$.} We set the display cost $b_{irj}$ to zero and $\gamma=1$ so we can work with the \emph{CTR} directly as the payoffs (therefore, we interpret the cost of experimentation as the opportunity cost to the advertiser of not showing the best combination.) We simulate 1,000 values for the expected \emph{CTR}s of the 6 \emph{C-DA} combinations from uniform distributions (with supports shown in Figure (\ref{fig:simul-setup})). Under these values, $C_{1}$-$DA_{1}$ has the highest expected \emph{CTR} amongst the \emph{C-DA} combinations, and $C_{1}$-$TA_{1}$ the highest amongst the \emph{C-TA} combinations. We run the TS for each simulated value to obtain 1,000 bandit replications. For each replication, we update probabilities over batches of 100 observations, and stop the sampling when we have 1000 batches of data. Then, we report in Figure (\ref{fig:valid}), box-plots across replications of the performance of the TS as batches of data are collected, plotting these at every $10^{\textrm{th}}$ batch.

\subsection{Algorithm Performance}

\begin{figure}[htbp]

\begin{minipage}{.6\linewidth}
\centering
\subfloat[Unit-free Regret]{\label{valid:a}\includegraphics[width=\textwidth]{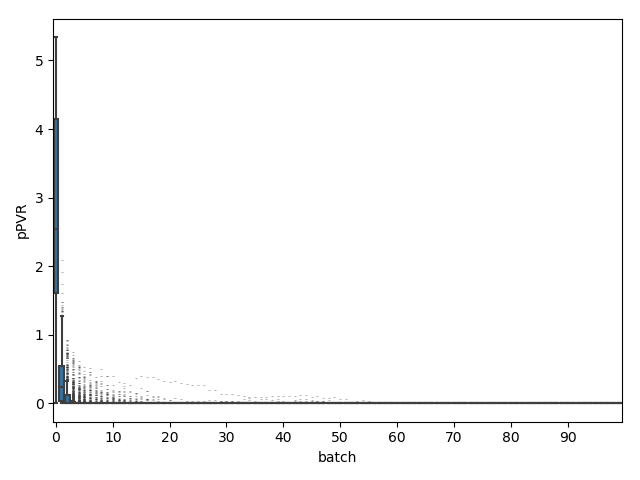}}
\end{minipage}
\begin{minipage}{.6\linewidth}
\centering
\subfloat[Expected Regret]{\label{valid:b}\includegraphics[width=\textwidth]{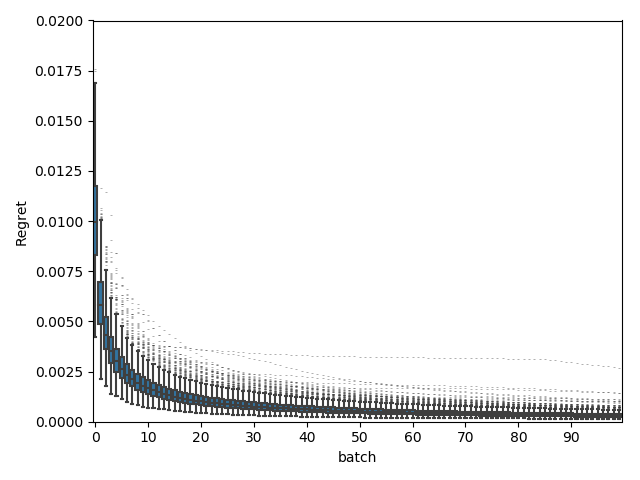}}
\end{minipage}\par
\centering
\subfloat[Pr(True-Best \emph{C-TA} Combination is Current-Best) ]{\label{valid:c}\includegraphics[width=0.6\textwidth]{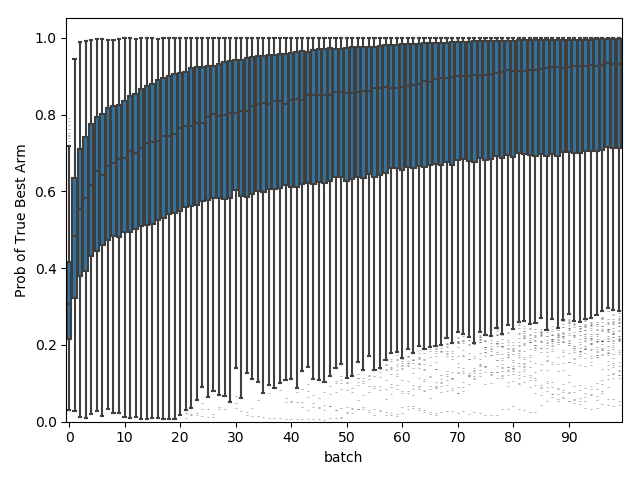}}

\caption{Results from 1,000 Replications}
\label{fig:valid}
\end{figure}

Figures (\ref{valid:a} and \ref{valid:b}) plot the evolution over batches in the unit-free regret (\textit{pPVR}) and the expected regret per impression, where the latter is defined as the expected clicks lost per impression in a batch when displaying a creative other than the true-best for each \textit{DA}, evaluated at the true parameters.\footnote{Specifically, the expected regret per impression in each batch $t$ is $\sum_{k\in\mathbb{K}}\sum_{j\in O(k)}\hat{p}(j|k)\sum_{r\in\mathbb{R}}w_{rjt}(\theta_{rj}^{\textrm{true}}-\underset{r\in\mathbb{R}}{\max}\theta_{rj}^{\textrm{true}})$.}  If the TS progressively allocates more traffic to creatives with higher probability of being the best arm in each context (\emph{DA}), the regret should fall as more data is accumulated. Consistent with this, both metrics are seen to fall as the number of batches increases in our simulation. The cutoff of 0.01 \textit{pPVR} is met in 1,000 batches in all replications. Figure (\ref{valid:c}) shows the posterior probability implied by TS in each batch that the true-best \emph{C-TA} is currently the best.\footnote{Note, these probabilities are not the same as the distribution of traffic allocated by the TS, since traffic is allocated based on \textit{DA} and not \textit{TA}.} The posterior puts more mass on the true-best combination as more batches are sampled. These results establish the face validity of the algorithm as a viable way of finding the best \emph{C-TA} combination in this setting, while minimizing regret.

\begin{figure}[htbp]
\begin{minipage}{.5\linewidth}
\centering
\subfloat[Total Regret at Stopping]{\label{Fig:tsea1}\includegraphics[width=\textwidth]{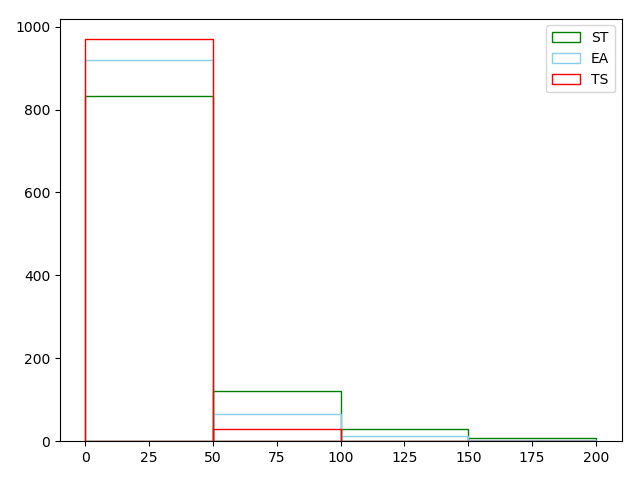}}
\end{minipage}
\begin{minipage}{.5\linewidth}
\centering
\subfloat[Sample Size at Stopping]{\label{Fig:tsea2}\includegraphics[width=\textwidth]{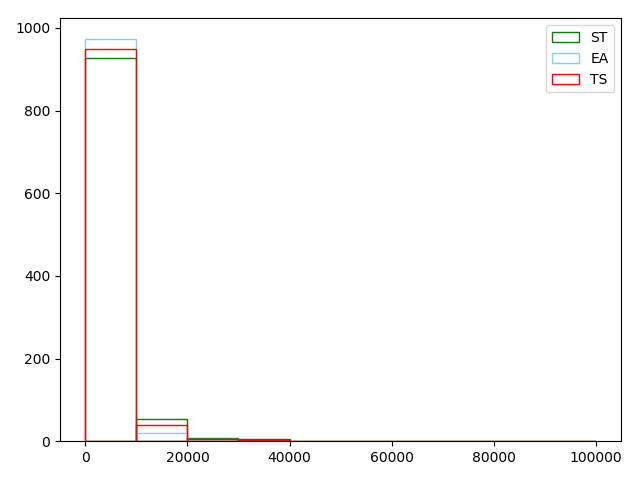}}
\end{minipage}\par

\centering
\subfloat[Pr(True-Best \emph{C-TA} Combination is Best at Stopping)]{\label{Fig:tsea3}\includegraphics[width=.5\textwidth]{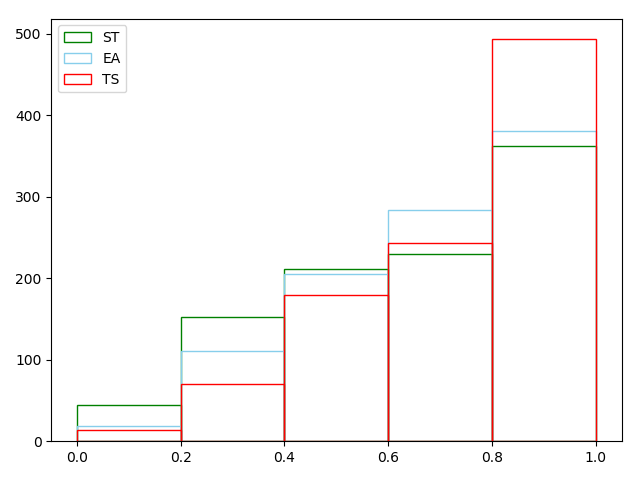}}

\caption{TS vs. Equal Allocation and Split-Testing}
\label{fig:ts}
\end{figure}

Figure (\ref{fig:ts}) now compares the proposed TS algorithm to an Equal Allocation algorithm (henceforth ``EA'') and a Split-Testing algorithm (henceforth ``ST''). EA is analogous to ``A/B/n'' testing in that it is non-adaptive: the allocation of traffic to creatives for each \textit{DA} is held fixed, and not changed across batches. Instead, in each batch, we allocate traffic equally to each of the $r\in\mathbb{R}$ creatives for each \emph{DA}. ST follows the design described in $\mathsection$\ref{sec:lit-review}, and traffic is allocated at the level of \emph{C-TA} (rather than \emph{C-DA}) combinations. Each user is assigned randomly with fixed, equal probability to one of $R\times K$ \emph{C-TA} arms (4 in this simulation), and a creative is displayed only if a user's features match the arm's \emph{TA} definition. 

To do the comparison, we repeat the same 1,000 replications as above with the same configurations, but this time stop each replication when the criterion in equation (\ref{eq:pPVR}) is reached. In other words, for each of TS, EA and ST algorithms, we maintain a posterior belief about the best \emph{C-TA} combination, which we update after every batch.\footnote{Note that, we do not need to partition the \emph{TA}s under ST, and instead directly set up the model at the \emph{C-TA} level under ST.} In TS, the traffic allocation reflects this posterior adaptively, while in EA and ST, the traffic splits are held fixed; and the same stopping criteria is imposed in both. All parameters are held the same.

Figure (\ref{Fig:tsea1}) shows that TS generates the smallest amount of expected regret, and the sample sizes required to exit the experiments under TS are between those under EA and those under ST (Figure  (\ref{Fig:tsea2})). This is because the expected regret per impression under EA and ST remains constant over batches, while as Figure (\ref{valid:b}) demonstrated, the expected regret per impression under TS steadily decreases as more batches arrive. ST generates the most regret and requires the largest sample sizes, since it is not only non-adaptive, but also discards a portion of the traffic and the information that could have been gained from this portion. Figure \ref{Fig:tsea3} shows that the TS puts more mass at stopping on the true-best \emph{C-TA} combination compared to EA and ST. Across replications, this allows TS to correctly identify the true-best combination 85.8\% of the time at stopping, compared to 77.8\% for EA and 70.8\% for ST. Overall, the superior performance of the TS relative to EA are consistent with the experiments reported in \cite{scott2010}.

\subsection{Degree of Overlap among Target Audiences}\label{subsec:overlap-exps}

The next set of experiments assesses the extent to which the degree of audience overlap affects the performance of the proposed TS algorithm. We use simulations to demonstrate the cross-audience learning effect in the algorithm, and to explore how it balances the effects of increased payoff similarity between the \emph{TA}s on performance. From a practical perspective, this simulation helps assess circumstances under which the sampler can reliably learn the best \emph{C-TA} combination (thereby representing an attractive scenario for the platform to run the test), versus not (and thereby representing an unattractive scenario for the platform to run the test).

\begin{figure}[!htbp]

\begin{minipage}{.5\linewidth}
\centering
\subfloat[Sample Size]{\label{ol_1:c}\includegraphics[width=\textwidth]{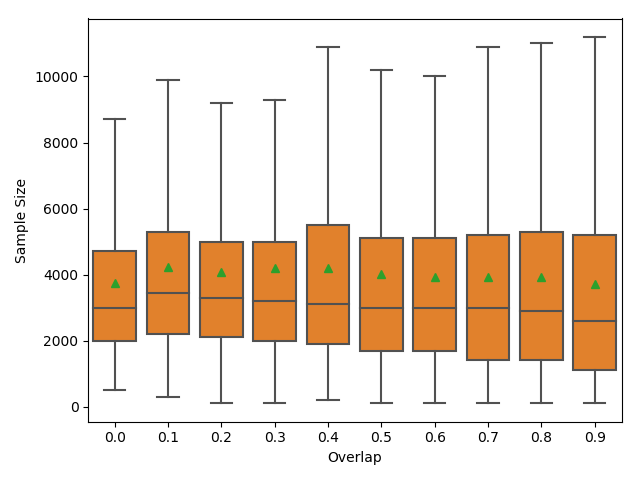}}
\end{minipage}
\begin{minipage}{.5\linewidth}
\centering
\subfloat[Total Exp. Regret]{\label{ol_1:d}\includegraphics[width=\textwidth]{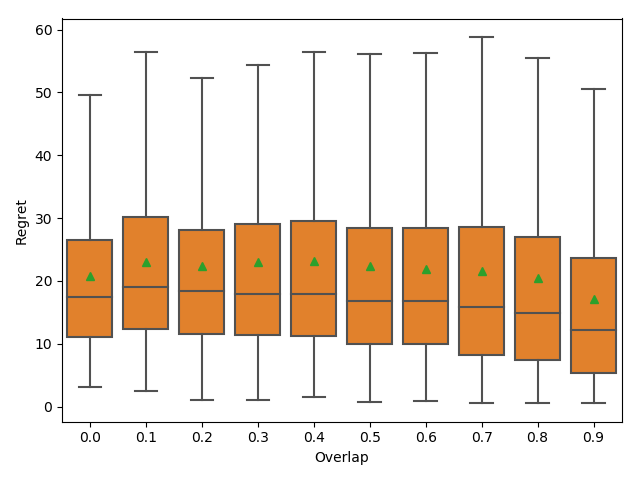}}
\end{minipage}\par

\begin{minipage}{.5\linewidth}
\centering
\subfloat[Prop. of True-Best Found]{\label{ol_1:a}\includegraphics[width=\textwidth]{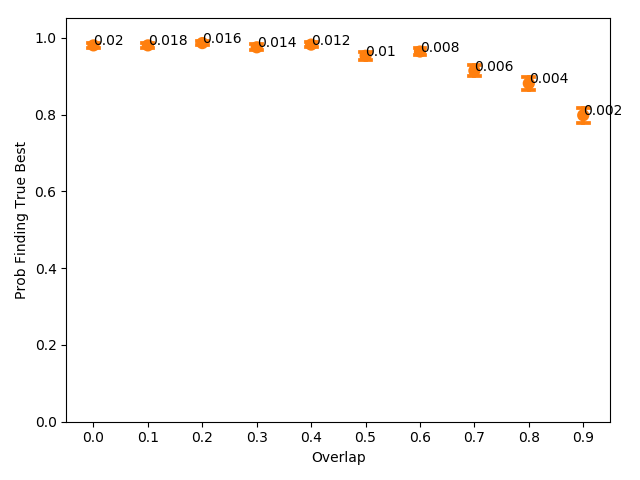}}
\end{minipage}
\begin{minipage}{.5\linewidth}
\centering
\subfloat[Post. Prob of Best at Stop]{\label{ol_1:b}\includegraphics[width=\textwidth]{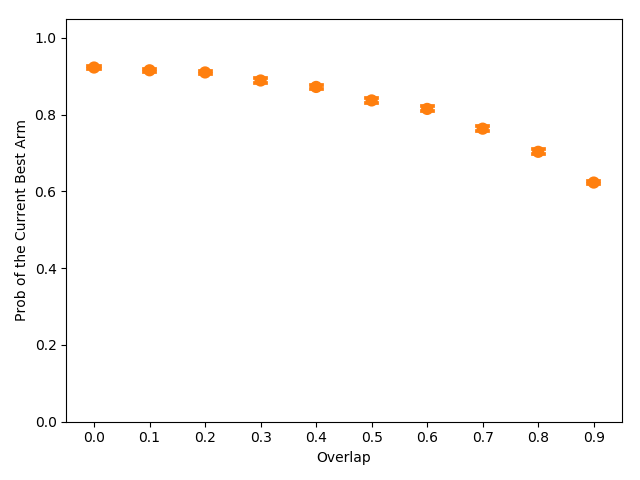}}
\end{minipage}\par

\caption{TS Performance with Increasing Overlap}
\label{fig:ol_1}
\end{figure}

We first fix the \emph{CTR}s of the six \emph{C-DA} combinations $C_{1}$-$DA_{1}$, $C_{2}$-$DA_{1}$, $C_{1}$-$DA_{2}$, $C_{2}$-$DA_{2}$, $C_{1}$-$DA_{3}$, $C_{2}$-$DA_{3}$ to be [.01,.03,.03,.05,.025,.035]. We vary the size of the overlapped audience, i.e. $\Pr\left(DA_{2}|TA_{1}\right)=\Pr\left(DA_{2}|TA_{2}\right)$, on a grid from $0$-$.9$. For each value on the grid, we run the TS for 1,000 replications, taking the  6 \emph{C-DA} \emph{CTR}s as the truth, stopping each replication according to equation (\ref{eq:pPVR}). We then present in Figure \ref{fig:ol_1} box-plots across these replications as a function of the degree of overlap. As the degree of overlap increases along the $x$-axis, the two target audiences become increasingly similar, increasing cross-audience learning, but decreasing their payoff differences.

Figures (\ref{ol_1:c} and \ref{ol_1:d}) show that sample sizes required for stopping and total expected regret per impression remain roughly the same as overlap increases, suggesting the two effects largely cancel each other. \footnote{Figure \ref{ol_1:d} suggests a possible decrease in total expected regret with increased overlap. This may be caused by a feature of the simulation setup that the overlapping \emph{DA} (i.e., \emph{$DA_{2}$}) has smaller payoff difference between the 2 \emph{CA}s than the non-overlapping \emph{DA}s. As the degree of overlap increases, the overlapping part dominates the non-overlapping part, making the regret smaller. If we impose the same payoff difference across all \emph{DA}s, we find this decline disappears.}

Figure (\ref{ol_1:a}) shows the proportion of 1,000 replications that correctly identify the true-best \emph{C-TA} combination as the best at stopping. The annotations label the payoff difference in the top-2 combinations, showing that the payoffs also become tighter as the overlapping increases. We see that the TS works well for reasonably high values of overlap, but as the payoff differences become extremely small, it becomes increasingly difficult to correctly identify the true-best \emph{C-TA} combination. Figure (\ref{ol_1:b}) explains this pattern by showing that the posterior probability of the best combination identified at stopping also decreases as the payoff differences grow very small. Finally, the appendix presents additional experiments that show that the observed degradation in performance of the TS at very high values of overlap disappears in a pure cross-audience learning setting.

Overall, these simulations suggest that the proposed TS is viable in identifying best \emph{C-TA} combinations for reasonably high levels of \emph{TA} overlap. The TS does this by leveraging cross-audience learning. If the sampler is to be used in situations with extreme overlap, it may be necessary to impose additional conditions on the stopping rule based on posterior probabilities, in addition to the ones based on $pPVR$ across contexts in equation (\ref{eq:pPVR}). This is left for future research.

\section{Deployment}
We designed an experimentation product based on the proposed TS algorithm.  The goal of the product is to help advertisers in \texttt{JD.com}'s marketplace improve their digital ad campaigns by discovering good target audience and creative combinations. 

To use the product, an advertiser starts by setting up a test ad-campaign on the product system. The test campaign is similar to a typical ad-campaign, involving advertiser-specified rules for bidding, budget, duration etc. The difference is that the advertiser defines $K$ \emph{TA}s  and binds $R$ creatives to the test-campaign, rather than one as typical; and the allocation of creatives to a user impression is managed by the TS algorithm. Both $K$ and $R$ are limited to a max of 5 so as to restrict the number of parameters to learn in the test. Because the algorithm disjoints \emph{TA}s, the number of contexts grows combinatorially as $K$ increases, and this restriction keeps the total \emph{C-TA} test combinations manageable.

When a user arrives at \texttt{JD.com}, the ad-serving system retrieves the user's characteristics. If the characteristics activate the tag(s) of any of the $K$ \emph{TA}s, and satisfies the campaign's other requirements, the TS chooses a test creative according to the adaptively determined probability, and places a bid for it into the platform's auction system. The bids are chosen by the advertiser, but are required to be the same for all creatives in order to keep the comparison fair.  The auction includes other advertisers who compete to display their creatives to this user. The system collects data on the outcome of the winning auctions and whether the user clicks on the creative when served; updates parameters every 10 minutes; and repeats this process until the stopping criterion is met and the test is stopped. The data are then aggregated and relevant statistical results regarding all the \emph{C-TA} combinations are delivered to the advertiser. See \texttt{https://jzt.jd.com /gw/dissert/jzt-split/1897.html} for a product overview.

The next sub-section presents a case-study based on one test run on the product. Though many of the other tests ran on the product platform exhibit similar patterns, there is no claim this case-study is representative: we picked it so it best illustrates for the reader some features of the test environment and the performance of the TS. 

\subsection{Case-Study}
\begin{figure}

\includegraphics[width=0.75\textwidth]{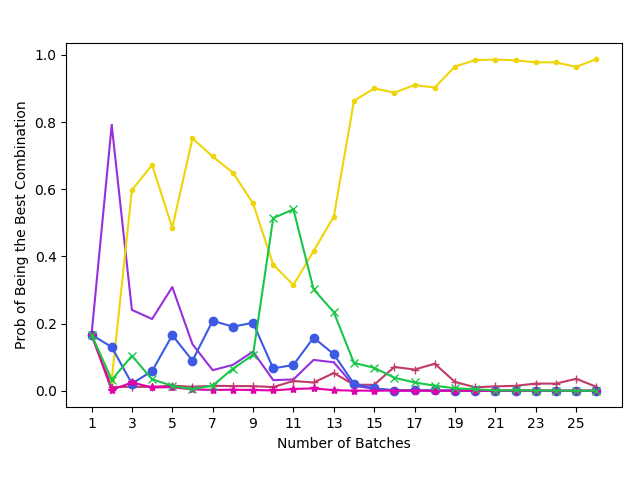}
\centering
\caption{Results from Practical Implementation}
\label{fig:exp}
\end{figure}

The case-study involves a large cellphone manufacturer. The advertiser defined 2 \emph{TA}s and 3 creatives. The 2 \emph{TA}s overlap, resulting in 3 \emph{DA}s. Figure (\ref{fig:exp}) shows the probability that each \emph{C-TA} combination is estimated to be the best as the test progresses. The 6 possible combinations are shown in different colors and markers. During the initial 12 batches (2 hours), the algorithm identifies the ``*" and ``+" combinations to be inferior and focuses on exploring the other 4 combinations. Then, the yellow ``." combination starts to dominate the other combinations until the test stops. When the test ends, this combination is chosen as the best. The advantage of the adaptive aspect is that most of the traffic during the test is allocated to this combination (see $y$-axis), so that the advertiser does not unnecessarily waste resources on assessing combinations that were learned to be inferior early on.

The experiment lasted a bit more than 6 hours with a total of 18,499 users and 631 clicks. The estimated \emph{CTR}s of the six \emph{C-TA} combinations $C_{1}$-$TA_{1}$, $C_{2}$-$TA_{1}$, $C_{3}$-$TA_{1}$ (yellow ``." combination), $C_{1}$-$TA_{2}$, $C_{2}$-$TA_{2}$, $C_{3}$-$TA_{2}$ at stopping are [.028,.034,.048,.028, .017,.036]. Despite the short time span, the posterior probability induced by the sampling on the yellow ``." combination being the best is quite high (98.4\%).

We use a back-of-the-envelope calculation to assess the economic efficiency of TS relative to EA in this test. We use the data to simulate a scenario where we equally allocate across the creatives the same amount of traffic as this test used. We find TS generates 52 more clicks (8.2\% of total clicks) than EA.

The quick identification of the best arm in this test may be due to the relatively large differences in the \emph{CTR}s across different combinations. The difference in the top-2 combinations is around 1\% for each of the \emph{TA}s and across all combinations. As we suggested in $\mathsection$\ref{subsec:overlap-exps}, larger differences in the payoffs may result in shorter test span and higher posterior probabilities on the best combinations. In other tests, we found the product performs well even in situations where the creatives are quite similar and $K,R$ are close to $5$, without requiring unreasonable amounts of data or test time so as to make it unviable. Scaling the product to allow for larger sets of test combinations is a task for future research and development.

\section{Conclusion}
An adaptive algorithm to identify the best combination among a set of advertising creatives and \textit{TA}s is presented. The novel aspect of the algorithm is to accommodate the possibility of overlap in the \textit{TA}s, which is a pervasive feature of digital advertising settings.  Overlap in the \textit{TA}s makes it difficult to sort between the relative attractiveness of various audiences. The proposed method addresses this issue, while adapting the allocation of traffic during the test to what is learned so as to minimize advertiser regret. Experiments show that the proposed method is more efficient compared to naive ``split-testing'' or non-adaptive ``A/B/n'' testing based methods. The approach assumes that creatives do not induce long-term dependencies, for instance, that they do not affect future user arrival rates, and that auctions are unrelated to each other, for instance due to the existence of a binding budget constraint. These assumptions justify framing the problem as a multi-armed bandit, and could be relaxed in future work, by using a more general reinforcement learning framework.


\bibliography{MAB.bib}
\bibliographystyle{acm}

\appendix
\renewcommand{\thesubsection}{\Alph{subsection}}
\numberwithin{equation}{subsection}

\section*{Appendix}

\subsection{Simulation: Increasing Overlap with Pure Cross-Audience Learning}

This simulation is set-up to demonstrate the \emph{cross-audience learning} effect induced by increasing overlap in \emph{TA}s. The idea of the simulation is to explore variation in performance with overlap, while holding fixed the payoff difference between the compared \emph{TA}s. Consider a similar setup with 2 creatives and 2 overlapping \emph{TA}s as before. We fix the \emph{CTR}s of the overlapped audience, $C_{1}$-$DA_{2}$, $C_{2}$-$DA_{2}$ to be [.015, .025] and the \emph{CTR}s of the four \emph{C-TA} combinations $C_{1}$-$TA_{1}$, $C_{2}$-$TA_{1}$,$C_{1}$-$TA_{2}$, $C_{1}$-$TA_{2}$ to be [.035, .05, .015, .03]. We vary the size of the overlapped audience, i.e. $\Pr\left(DA_{2}|TA_{1}\right)=\Pr\left(DA_{2}|TA_{2}\right)$, on a grid from $0$-$.9$. For each value on the grid, we run the TS for 1,000 replications, stopping each according to equation (\ref{eq:pPVR}).\footnote{That is, for each value of overlap on the grid, we compute the \emph{CTR}s for $C_{1}$-$DA_{1}$, $C_{2}$-$DA_{1}$,$C_{1}$-$DA_{3}$, $C_{1}$-$DA_{3}$  that generate the 2 fixed \emph{C-DA} and 4 \emph{C-TA} \emph{CTR} values above, and run 1,000 replications of the TS taking the  6 \emph{C-DA} \emph{CTR}s as the true-values.} Since the payoffs of the \emph{C-TA} combinations remain the same as the overlap changes, this helps isolate the effects of cross-audience learning. 

\begin{figure}[!htbp]

\begin{minipage}{.5\linewidth}
\centering
\subfloat[Sample Size]{\label{ol_2:c}\includegraphics[width=\textwidth]{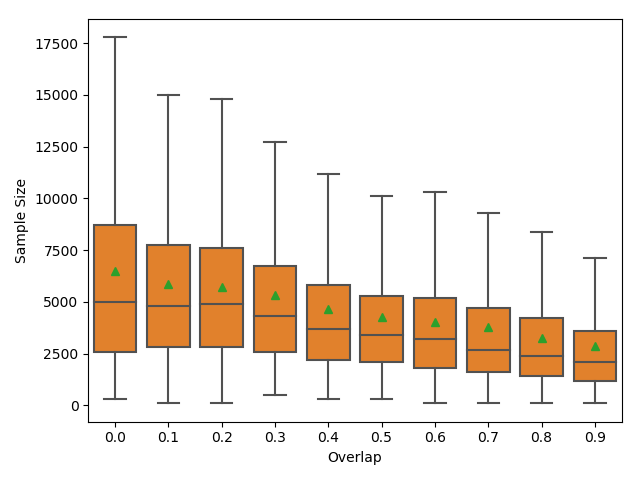}}
\end{minipage}
\begin{minipage}{.5\linewidth}
\centering
\subfloat[Total Exp. Regret]{\label{ol_2:d}\includegraphics[width=\textwidth]{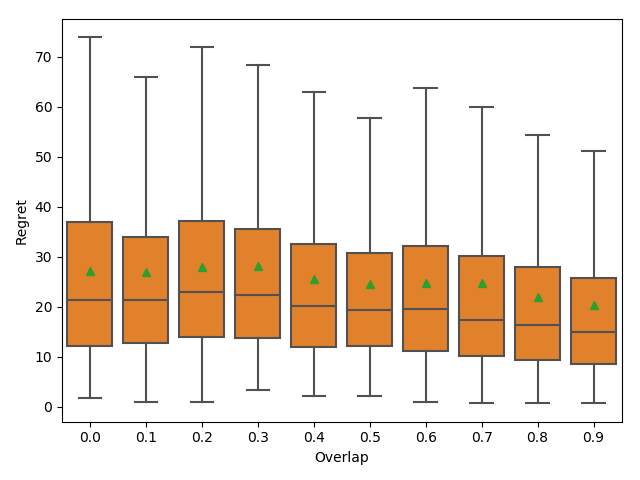}}
\end{minipage}\par

\begin{minipage}{.5\linewidth}
\centering
\subfloat[Prop. of True-Best Found]{\label{ol_2:a}\includegraphics[width=\textwidth]{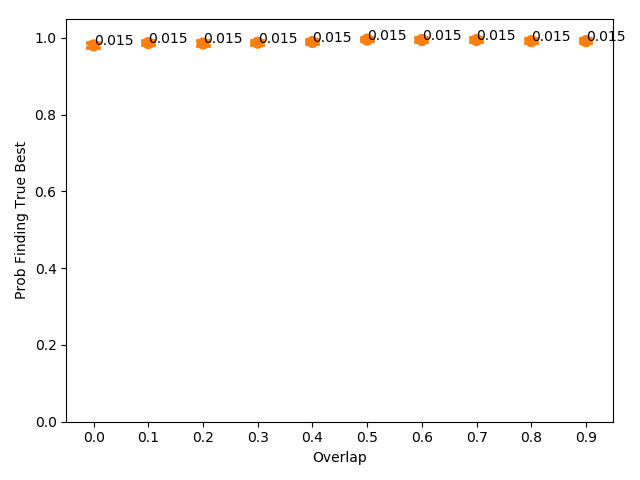}}
\end{minipage}
\begin{minipage}{.5\linewidth}
\centering
\subfloat[Post. Prob of Best at Stop]{\label{ol_2:b}\includegraphics[width=\textwidth]{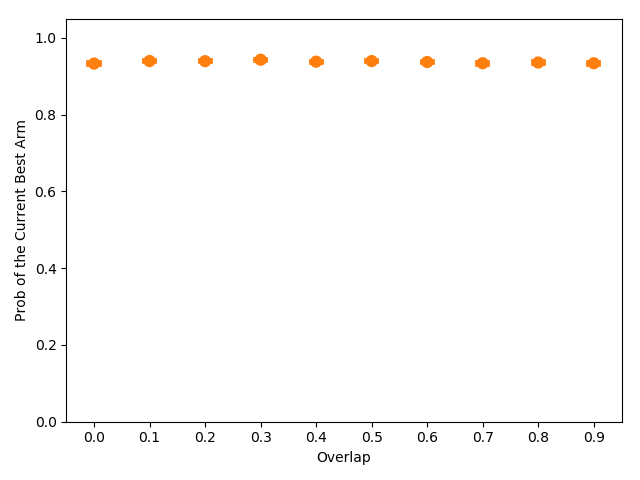}}
\end{minipage}\par

\caption{TS Performance with Increasing Overlap but Fixed Payoff Differences}
\label{fig:ol_2}
\end{figure}

Figure \ref{fig:ol_2} shows box-plots across replications as a function of the degree of overlap. Reflecting the cross-audience learning, the sample sizes decrease steadily as the overlap increases (Figure (\ref{ol_2:c})). Expected regret per impression may increase as overlap increases because the payoff difference between the non-overlapping \emph{DA}s increases with overlap; or it may fall because of faster learning. Figure (\ref{ol_2:d}) shows the net effect is somewhat negative and expected regret declines as  overlap increases. Figure (\ref{ol_2:a}) shows the proportion of replications that correctly identify the true-best \emph{C-TA} combination as the best at the end of each replication. We see the proportion of correctly identified combinations remain high as the degree of overlap increases. Finally, Figure (\ref{ol_2:b}) shows there is no degradation in performance in terms of the posterior probability accumulated on the best combination at stopping. 

Overall, these show that under a pure \emph{cross-audience learning} scenario, increased overlap between the \emph{TA}s does not degrade performance of the sampler in a meaningful way.

\end{document}